\documentclass[10pt,twocolumn,letterpaper]{article}

\usepackage{cvpr}              %

\usepackage{graphicx}
\usepackage{amsmath}
\usepackage{amssymb}
\usepackage{pifont} %
\usepackage{enumitem}
\usepackage{booktabs}
\usepackage{setspace}
\usepackage[misc]{ifsym}
\newcommand{\cmark}{\ding{51}}
\newcommand{\xmark}{\ding{55}}
\usepackage[dvipsnames]{xcolor}

\definecolor{cvprblue}{rgb}{0.21,0.49,0.74}
\usepackage[pagebackref,breaklinks,colorlinks,citecolor=cvprblue]{hyperref}

\usepackage{textpos}  %

\newcommand{\refsec}[1]{Sec.~\ref{sec:#1}}

\newcommand{\reftbl}[1]{Tab.~\ref{tab:#1}}
\newcommand{\reffig}[1]{Fig.~\ref{fig:#1}}

\newcommand{\refeq}[1]{Eq.~\ref{eq:#1}}

\newcommand{\lblfig}[1]{\label{fig:#1}}
\newcommand{\lblsec}[1]{\label{sec:#1}}
\newcommand{\lbleq}[1]{\label{eq:#1}}

\newcommand{\lbltbl}[1]{\label{tab:#1}}

\newcommand{\tablestyle}[2]{\setlength{\tabcolsep}{#1}\renewcommand{\arraystretch}{#2}\centering\small}
\newlength\savewidth\newcommand\shline{\noalign{\global\savewidth\arrayrulewidth
  \global\arrayrulewidth 1pt}\hline\noalign{\global\arrayrulewidth\savewidth}}

\newcommand{\customfootnotetext}[2]{{%
  \renewcommand{\thefootnote}{#1}%
  \footnotetext[0]{#2}}}%

\def\paperID{****} %
\def\confName{CVPR}
\def\confYear{2024}

\title{Pixel Aligned Language Models}

\author{Jiarui Xu$^{1,2*}$ \quad Xingyi Zhou$^{1}$ \quad Shen Yan$^{1}$ \quad Xiuye Gu$^{1}$ \\
Anurag Arnab$^{1}$ \quad Chen Sun$^{1}$ \quad Xiaolong Wang$^{2}$ \quad Cordelia Schmid$^{1}$\\
{$^1$Google Research \quad \quad $^2$UC San Diego}
}

\newcommand{\ourmethod}{PixelLLM\xspace}
\newcommand{\ourname}{Pixel-Aligned Language Model\xspace}

\begin{document}
\twocolumn[{%
\renewcommand\twocolumn[1][]{#1}%
\maketitle
\begin{center}
    \centering
    \captionsetup{type=figure}
    \includegraphics[width=0.9\textwidth]{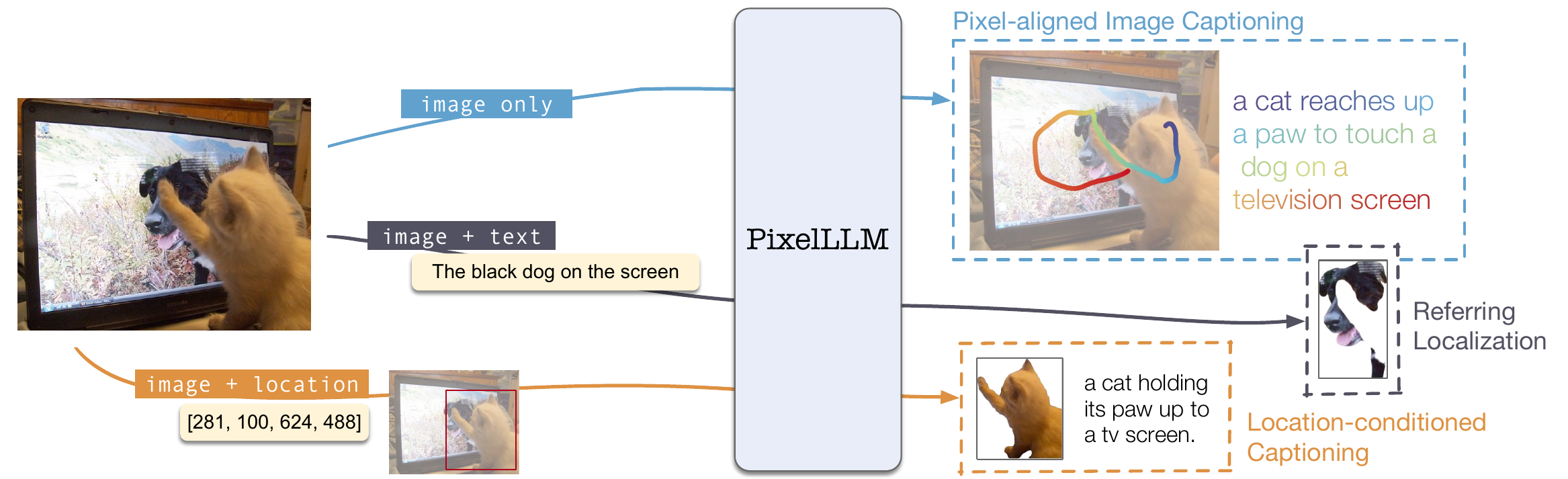}
    \caption{
        We propose \ourname{}~(\ourmethod) to equip large language models with localization capability.
        The model is pre-trained on localized image captioning data~\cite{pont2020connecting}, where each word is labeled with a pixel location, to learn the alignment between words and image pixels.
        \ourmethod{} can be applied to various localization tasks,
        for example,
        \textit{location-conditioned captioning} when taking location as input, and \textit{referring localization} when generating locations as outputs.
    }
    \lblfig{teaser}
\end{center}%
}]
\customfootnotetext{}{\!\!\!\!\!\!\!\!\!\!\!\!*Work done during a Google internship. \Letter\{jiaruixu, zhouxy\}\!@\!google.com\!\!}
\begin{textblock*}{.9\textwidth}[.5,0](0.5\textwidth, -0.45\textwidth)
    \centering
    \vspace{-5mm}
    {\url{https://jerryxu.net/PixelLLM/}}
\end{textblock*}
\vspace{-1.2em}
\begin{abstract}
\vspace{-0.5em}
Large language models have achieved great success in recent years, so as their variants in vision.
Existing vision-language models can describe images in natural languages, answer visual-related questions, or perform complex reasoning about the image.
However, it is yet unclear how localization tasks, such as word grounding or referring localization, can be performed using large language models.
In this work, we aim to develop a vision-language model that can take locations, for example, a set of points or boxes, as either inputs or outputs.
When taking locations as inputs, the model performs location-conditioned captioning,
which generates captions for the indicated object or region.
When generating locations as outputs,
our model regresses pixel coordinates for each output word generated by the language model,
and thus performs dense word grounding.
Our model is pre-trained on the Localized Narrative dataset, which contains pixel-word-aligned captioning from human attention.
We show our model can be applied to various location-aware vision-language tasks, including referring localization, location-conditioned captioning, and dense object captioning, archiving state-of-the-art performance on RefCOCO and Visual Genome.

\end{abstract}
\vspace{-1.5em}    
\section{Introduction}
\lblsec{intro}

Imagine a baby waving their hands, pointing to the colorful toys, and yelling the toy names.
Pointing and naming is a natural and convenient way to describe the visual world, and provide dense visual-language alignment that is synchronized over time.
Can we design an intelligent model~\cite{openai2022chatgpt,google2023bard,touvron2023llama,2020t5} that leverages such information to obtain
{vision-language alignment}?
Prior research~\cite{liu2023visual,zhu2023minigpt,li2023blip,wang2022git,li2023otter,gao2023llama, instructblip, bai2023qwen} has tried to align visual information to the pre-trained LLMs for various vision-language tasks such as producing long and detailed descriptions~\cite{liu2023visual} or conversation~\cite{li2023blip}.
However, most of these works take the entire image as input, and produce all outputs in text.
Detailed understanding of specific regions and objects, and their exact locations, is not well-studied in the context of LLMs.
We aim to propose a new architecture and a training procedure that can achieve the goal, which helps us answer the important research question: can large \emph{language} models achieve spatial understanding and reasoning from the \emph{visual} world?
If so, how?

To this end, we introduce \ourmethod{}, a vision-language model with fine-grained localization ability by
densely aligning each output word to a pixel location.
We realize this by adding a small MLP on top of the word features of a language model, to regress to a pixel location of each word.
The weights of the language model can be kept frozen, or updated by Low-rank finetuning (LoRA)~\cite{hu2021lora}.
Furthermore, our model can take the location prompt or text prompt, to generate outputs specific to the prompt.
\reffig{teaser} shows the tasks that our model can do or can adapt to.

\begin{figure*}[t]
	\center
	\includegraphics[width=0.95\linewidth]{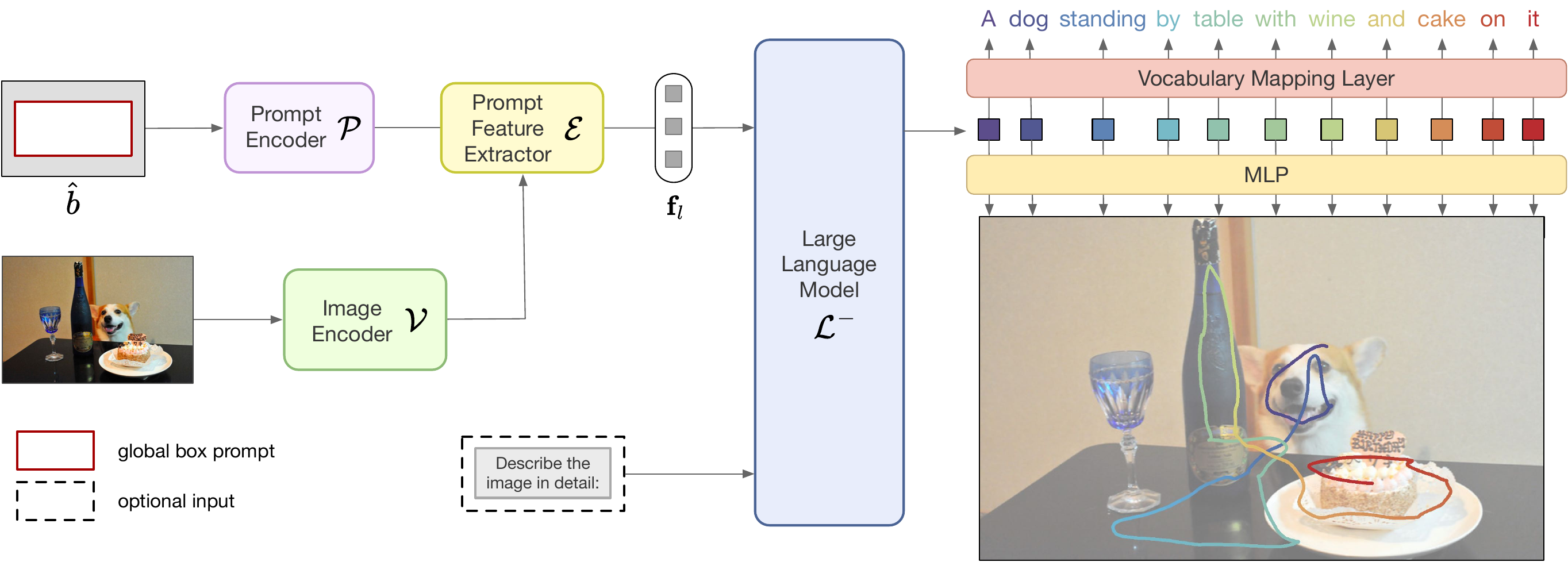}
        \vspace{-3mm}
	\caption{
		\label{fig:framework}
		\textbf{Overview of \ourmethod{} architecture for pixel-aligned captioning.} We first encode the input location prompt (global box prompt in this case) and the input image with the prompt encoder $\mathcal{P}$ and the image encoder $\mathcal{V}$ respectively. Then we input the prompt feature $\mathbf{l}$ and the image feature $\mathbf{f}$ into the prompt feature extractor to extract location-specific visual feature $\mathbf{f_l}$. The large language model $\mathcal{L}$ 
        then auto-regressively predicts the next text tokens conditioned on previous text tokens and the visual feature. We apply a simple MLP layer on the token features before the vocabulary mapping layer of LLM, which predicts the coordinates of each text token. The alignment between the caption and the trace is represented by color gradient \protect\includegraphics[width=8mm]{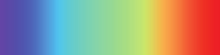}
  }
  \vspace{-3mm}
\end{figure*}

We show our concrete architecture in \reffig{framework}.
It includes an image encoder, a prompt encoder, and a prompt feature extractor which maps prompt-conditioned image features to the text embedding space.
These prompt-conditioned image features and an optional text prompt, are then directly fed as the prefix of a large-language model, which produces the captioning and per-word localization outputs.
We show this architecture is general and can adapt to various vision-language tasks, with any combination of language or location as input or output.

While our model requires dedicated word-pixel aligned data to train, we note 
that such annotations already exist at large scale~\cite{pont2020connecting}.
The localized narratives dataset~\cite{pont2020connecting} contains annotations of human annotators narrating a given image, together with a mouse trajectory of the annotators' attention during the narration.
This gives synchronized locations for all words in the narration sentence, which can be used to train our model.
While not all word-location pairs are visually meaningful or accurate,
we argue that they are valuable as they are from actual human attention.

We evaluate our model on popular vision tasks by adapting our architecture and fine-tuning on downstream datasets,
including referring localization on RefCOCO~\cite{yu2016modeling}, location conditioned captioning on RefCOCO and Visual Genome~\cite{krishna2017visual}, and dense object captioning on  Visual Genome~\cite{krishna2017visual}.
Our model achieves the state-of-the-art performance on all these tasks,
with $89.8$ P@0.5 on RefCOCO referring localization, $19.9$ CIDEr on Visual Genome conditioned captioning, and $17.0$ mAP on dense object captioning.
Ablations on RefCOCO~\cite{yu2016modeling}(\reftbl{ablation_localization}) show our dense per-pixel localization formulation is the key to the high performance, with a $3.7$ point gain compared to alternative localization formulations.

\noindent Our contributions are summarized below:
\begin{itemize}
\item We introduce PixelLLM, a vision-language model that outputs a caption for an input image together with the localization of each word. PixelLLM takes as input an image, and optionally a location prompt or a text prompt.
\item Our model can utilize the dedicated localized narrative dataset~\cite{pont2020connecting},
which comes with image captions and trajectories localizing each word, for per-word localization.
\item Our model is flexible and can be adapted to various vision-language tasks, including referring localization and segmentation, location-conditioned captioning, and dense captioning, with state-of-the-art performance.
\end{itemize}

\noindent Code and models will be released.

\section{Related work}
\lblsec{related-work}

\noindent\textbf{Large-scale vision and language models.}
Building large vision and language models that are generally capable of visual captioning or question answering is a trending research topic~\cite{wang2022git,radford2021learning,chen2022pali,wang2022image,instructblip, zhu2023minigpt, zhang2023llamaadapter, huang2023language}. 
Flamingo~\cite{alayrac2022flamingo} inserts visual features into language models through a gating function.
BLIP2~\cite{li2023blip} connects vision features~\cite{fang2023eva} to frozen large-language models~\cite{2020t5} using learnable queries~\cite{carion2020end}.
LLaVa~\cite{liu2023visual} distills large language models to visual inputs by creating instructional image-text pairs.
The majority of these works focus on language-only outputs.
While language is general enough for image-level tasks, region-level localization tasks are more challenging.
Our work builds on existing vision-language models~\cite{wang2022git,li2023blip},
and adds localization ability using a per-token regression head.

\noindent\textbf{Modeling locations in language models.}
We are not the first to study localization in the context of large language models~\cite{zhang2023gpt4roi,wang2023visionllm,lai2023lisa,chen2023shikra, you2023ferret}.
Pix2Seq~\cite{chen2021pix2seq} first proposes to model localization as spatial bin tokens in the vocabulary, and tackle object detection as an auto-regressive problem.
Following this idea, UniTab~\cite{yang2022unitab} embeds bounding boxes bin tokens in captions, which enables predicting object locations when seeing nouns in the output.
Kosmos~\cite{peng2023kosmos} uses a similar representation, and creates interleaved location-caption datasets by distilling grounding models~\cite{li2022grounded}.
A key distinction for our work is the localization representation.
While existing works try to model locations as a word in the language model vocabulary, we use a regression representation.
Also, most existing works~\cite{lai2023lisa,peng2023kosmos,liu2023visual} rely on distilled data from existing models~\cite{liu2023visual,wu2022grit}, and
our representation enables us to utilize the existing human annotation from Localized Narratives dataset~\cite{pont2020connecting}.

\noindent\textbf{Referring expression localization} is a popular task in language and vision, aiming to localize a sentence query~\cite{yu2016modeling, mao2016generation}.
MDETR~\cite{kamath2021mdetr} formulates this with the cross-attention mechanism in DETR~\cite{carion2020end} style, where the language features are the queries and the visual features are keys.
UniTab~\cite{yang2022unitab} integrates box output as the follow-up spatial tokens in a language decoder.
Shikra~\cite{chen2023shikra} directly extracts formatted bounding boxes from the raw string generated by the large language model.
We tackle this task by regression given the language and vision features using a large-language model.

\noindent\textbf{Attention trajectory modeling} is a new task enabled by the localized narratives dataset~\cite{pont2020connecting}.
It aims to align a long and detailed image caption to a trajectory of human attention, collected from the mouse trace while annotating the caption.
The only existing work that evaluates on this task is MITR~\cite{meng2021connecting}, which trains linear layers on frozen region features.
While this is effective, the knowledge learned in MITR couldn't transfer to other tasks given the task-specific design.
To the best of our knowledge, we are the first to propose an end-to-end training framework for this task and train it at scale.

\section{Preliminary}

\lblsec{prelim}

Given an input image $\mathbf{I} \in \mathbb{R}^{H \times W \times 3}$, our first goal is to generate an image caption $\mathbf{s}$, 
a sequence of word tokens\footnote{Technically, a word can be multiple tokens. For presentation clarity, we assume a word is a token. We use ``word'' and ``token'' interchangeably.}
:\ $\mathbf{s} = [w_1, w_2, \cdots, w_n]$. Each token $w_i \in [0, |V|]$ is an integer indexing a vocabulary $V$.
This task is known as image captioning~\cite{chen2015microsoft}, and is widely studied in vision and language research~\cite{wang2022git,li2023blip,chen2022pali}.

Optionally, we hope to take additional prompt inputs, to specify the region or concepts that the model should focus on.
The prompts can be locations, indicated by a sequence of points $\mathbf{x} \in \mathbb{R}^{m \times 2}$ or a box $b \in \mathbb{R}^4$, or texts (in the form of a sequence of word tokens $\mathbf{t} = [w_1, \cdots]$).
Before introducing our proposed approach, we review existing techniques for standard image captioning and input prompt encoding.

\noindent\textbf{Image captioning} produces a sentence $\mathbf{s}$ given an input image $\mathbf{I}$.
Popular architectures~\cite{wang2022git,chen2022pali,li2023blip} first encode the image as a feature $\mathbf{f} \in \mathbb{R}^{N \times C}$
using an image encoder $\mathcal{V}$ (\eg, a ViT~\cite{dosovitskiy2020image}),
where $N$ is the number of tokens and $C$ is the dimension size.
They then feed $\mathbf{f}$ to an auto-regressive~\cite{vaswani2017attention,graves2013generating} language model $\mathcal{L}$.
The language model produces the word sequence one-by-one, conditioning on the vision feature and previously predicted words: $w_i = \mathcal{L}(\textbf{f}, \mathbf{w}_{1:i-1})$.
The architecture of the language model can be a decoder-only architecture with a stack of self-attention layers~\cite{wang2022git}, or a pair of encoder-decoder with cross-attention~\cite{2020t5}.

\noindent\textbf{Prompt encoder}.
In addition to an image input, prompt encoders adapt other input modalities to the same feature space as the image feature $\mathbf{f}$.
We follow SAM~\cite{kirillov2023segment} to encode points or box coordinates using a location prompt encoder $\mathcal{P}$ consisting of a sine-cosine position embedding followed by linear layers. 
For text prompt, we follow GIT~\cite{wang2022git} and BLIP2~\cite{li2023blip} to concatenate query word embedding with visual features as the prefix features for the language model.

With these components, one can already compose a prompt-conditioned image captioning model.
However, it is not yet feasible to generate location outputs, \eg, localizing the queried concept or grounding the answer to a region.
Next, we introduce our \ourname{} that seamlessly integrates per-token localization ability into vision-language models.

\section{Pixel-Aligned Language Model}

\lblsec{methods}

Localization has been studied in multiple forms in computer vision.
For instance, localizing all instances given a set of vocabularies (object detection~\cite{lin2014microsoft}),
localizing a natural language query (referring localization~\cite{yu2016modeling}), or
associating each entity in a sentence to a bounding box (entity grounding~\cite{plummer2015flickr30k}).

Under the framework of vision-language models, we propose to formulate localization as \textit{aligning each word in the output sentence to a pixel location}.
Specifically, besides the sentence output $\textbf{s}$, 
we also output a sequence of points $\mathbf{p}$ with the same length as the sentence, 
$\mathbf{p} = [p_1, p_2, \cdots, p_n]$, $p_i \in \mathbb{R}^2$, 
each corresponds to the respective word token in the sentence.
In contrast to existing works~\cite{li2022grounded,yang2022unitab} that
only nouns may be grounded to a region,
we don't enforce the model to ignore non-visual tokens, so that the model can also learn relation terms, \eg ``holding''.

Next, we first introduce novel components of our architecture (\refsec{architecture}).
We then describe how we train it on large-scale densely annotated
datasets (\refsec{training}). 
Finally, we show our architecture can be applied to various localization tasks with no or minor modifications~(\refsec{tasks}).

\subsection{Architecture}
\lblsec{architecture}

\reffig{framework} gives an overview of our architecture.
The input is an image $\mathbf{I}$ and an optional location prompt $b$.
If no location prompt is provided, we use a global box prompt of the entire image, \ie $b = (0, 0, H, W)$, where $H$ and $W$ are the image size.
Otherwise, the model is expected to focus on the locations indicated by the prompt.
The output is the sentence $\mathbf{s}$ and its aligned point trajectory $\mathbf{p}$.

Using the image encoder $\mathcal{V}$ and location prompt encoder $\mathcal{P}$ introduced above, we obtain the image feature $\mathbf{f}=\mathcal{V}(\mathbf{I})$ and location prompt feature $\mathcal{P}(b)$.
Here $\mathbf{f}$ is the feature of the entire image,
we use an prompt feature extractor $\mathcal{E}$ to extract the feature specified by the location prompt:
\begin{equation}
\mathbf{f_l} = \mathcal{E}(\mathbf{f}, \mathcal{P}(b))
\end{equation}
where $\mathbf{f}_l \in \mathbb{R}^{N \times C}$ is the location-specific visual feature.

We instantiate the prompt feature extractor $\mathcal{E}$ using a two-way transformer with a set of learnable tokens $\mathbf{q}$, inspired by QFormer~\cite{li2023blip}.
Specifically, the two-way transformer takes 
$[\mathcal{P}(b), \mathbf{q}]$ and $\mathbf{f}$ as the query or key/value alternatively at each layer, and finally takes the learnable token features at the last layer.
The output feature $\mathbf{f}_l$ conveys the feature specific to the location prompt $b$. 
Conceptually, prompt feature extractor has a similar function as ROIAlign~\cite{he2017mask}, but it is learnable and doesn't require feature interpolation and sampling.
We compare and discuss the differences in \refsec{densecap}.

\noindent\textbf{Dense location outputs from language models.}
Given the location-specific feature $\mathbf{f}_l$, we can already feed it to a language model to do captioning, using auto-regressive decoding:
$w_i = \mathcal{L}(\mathbf{f}_l, \mathbf{w}_{1:i-1})$.
We note the last linear layer of a language model is a vocabulary mapping layer, mapping from the language feature space to the one-hot vocabulary index.
Let $\mathcal{L}^-$ denote the language model without the last vocabulary mapping layer, the decoding process (for simplicity, we show greedy decoding here) can be rewritten as
\begin{equation}
w_i = \text{argmax} (\mathbf{v} \cdot \mathcal{L}^-(\mathbf{f}_l, \mathbf{w}_{1:i-1}))
\lbleq{location-prediction}
\end{equation}
where $\mathbf{v} \in \mathbb{R}^{|V| \times C}$ is the weight of the linear vocabulary mapping layer.

To use the same language features for localization, we simply add a small MLP in parallel with the vocabulary mapping layer, which maps the language feature to a 2-dimension location output\footnote{There are multiple trace points per word token, we show the single point prediction for simplicity. In practice, we output the two points that tightly bound the trace of each token.}:
\begin{equation}
p_i = \text{MLP}(\mathcal{L}^-(\mathbf{f}_l, \mathbf{w}_{1:i-1}))
\lbleq{location-output}
\end{equation}
Note we do not feed back the localization output for auto-regressive decoding, to avoid affecting the original text decoding process.
The location prediction runs on the fly together with language decoding, only adding minor computation overhead.
This design is language model agnostic, and can be applied to any language model without interfering with the original language generation ability.
To take text prompt as input, we directly concatenate text prompt word embeddings with the visual feature $\mathbf{f_l}$.

\lblsec{dense-outputs}

\subsection{Training}
\lblsec{training}
We train our model using a human-annotated caption-location aligned dataset Localized Narrative~\cite{pont2020connecting} (LN).
Localized Narrative asks annotators to narrate a given image while simultaneously moving the mouse over the region they are describing.
The narration and the mouse trace are synchronized, which gives the location of each single word of the narration.
Although the mouse trace could be noisy, it is still a cheap and effective way to obtain dense location supervision.
Thus, this dataset contains all the needed triple annotations ($\mathbf{I}$, $\mathbf{s}$, $\mathbf{p}$): the image $\mathbf{I}$, the captioning sentence $\mathbf{s}$, and the location trajectory $\mathbf{p}$.

We use the standard label-smoothed cross-entropy loss to train the captioning output,
and use an $L_1$ regression loss to train the localization output:
\begin{equation}
L = \frac{1}{n}\sum_{i=1}^{n} (\text{CE}(\mathcal{L}(\mathbf{f}_l, \textbf{w}_{1:i-1}), {w}_i) + \lambda |\hat{p}_i - p_i|)
\lbleq{loss}
\end{equation}
where $\hat{p}_i$ is the predicted location for the $i$-th word, $\lambda$ is the localization loss weight, and $n$ is the caption length. 

\subsection{Adapting to downstream vision tasks.}
\lblsec{tasks}

Our architecture can take any combination of text/ location as inputs or outputs, and thus can be applied to various location-related vision tasks.
In this section, we show examples of adapting our architecture for three popular tasks: referring localization and segmentation, location-conditioned captioning, and dense object captioning.

\begin{figure}[t]
	\center
	\vspace{-1em}
	\includegraphics[width=\linewidth]{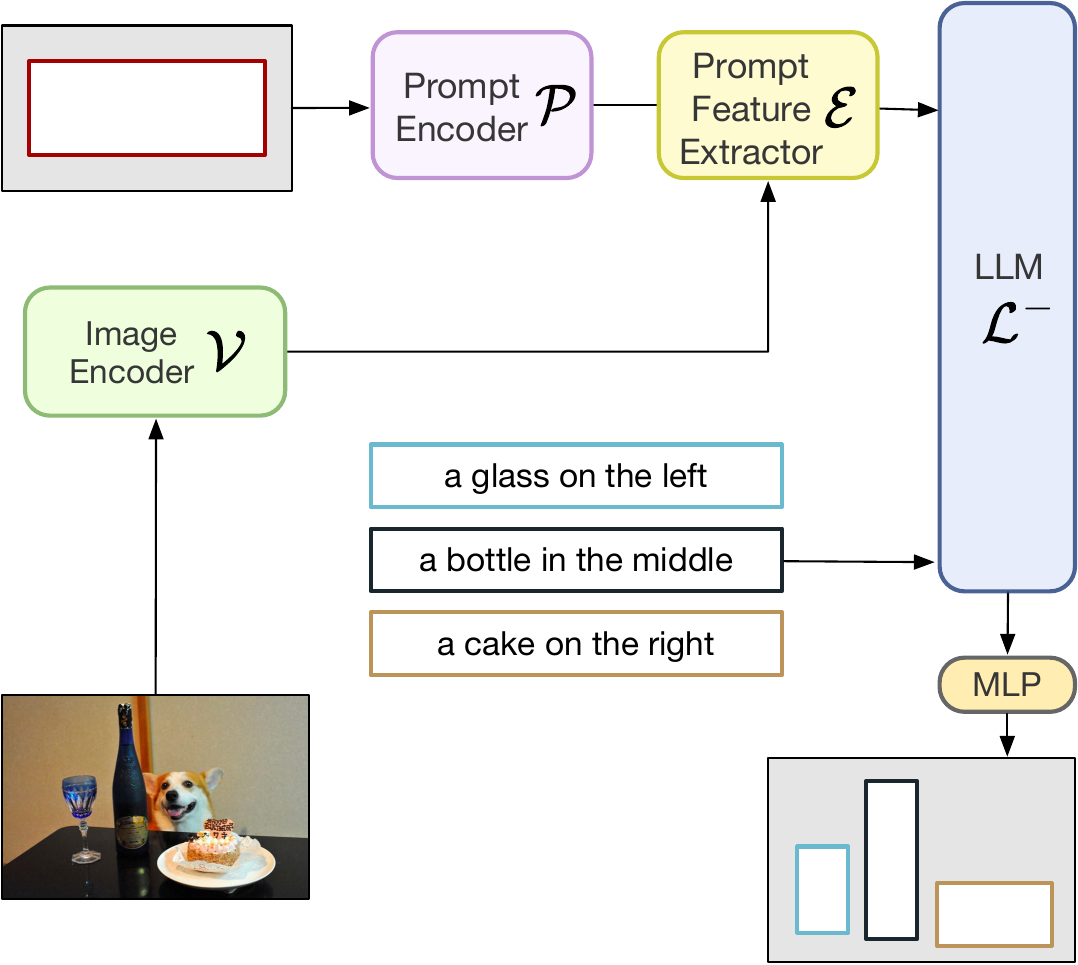}
	\vspace{-1em}
	\caption{
		\label{fig:refcoco}
		\textbf{Our model for referring expression localization pipeline.} To apply \ourmethod{}, we don't need to generate the text tokens.
  Instead, we directly input the query $\mathbf{t}$ into the LLM $\mathcal{L}^-$ to extract the token features before the vocabulary mapping layer.
        We then apply MLP to the last token predict the bounding boxes.
	}
	\lblfig{fig:refcoco}
 \vspace{-1em}
\end{figure}
\noindent\textbf{Referring localization and segmentation} takes an image $\mathbf{I}$ and a sentence query $\mathbf{t}$ as input, and aims to produce a bounding box $\hat{b} \in \mathbb{R}^4$ that corresponds to the query.
To apply our framework to this task, we set the location prompt as a global box (i.e., $\bar{b}=(0, 0, W, H)$), and use the query sentence as the conditioned sentence in \refeq{location-output}.
By default, our model outputs a trajectory, rather than a single bounding box.
While one can form a bounding box by taking the boundaries of the trajectory, we observe this is suboptimal, as the trajectory boundaries are not tight object boundaries required in evaluation.
We thus train the model to output an accurate object bounding box at the \texttt{<EOS>} token, using the same regression MLP layers.
We take the single box output at the \texttt{<EOS>} token as the output:
\begin{equation}
\hat{b} = \text{MLP}\Big(\mathcal{L}^-(\mathcal{E}(\mathcal{V}(\mathbf{I}), \mathcal{P}(\bar{b})), [\mathbf{t}, \texttt{<EOS>}])\Big)
\lbleq{referring}
\end{equation}

As our model already include the image backbone and prompt encoder from SAM~\cite{kirillov2023segment},
we can further obtain the segmentation mask by simply plugging in the mask decoder of SAM.
Our model can thus also used for referring segmentation, by producing a mask on top of the predicted box $\hat{b}$.

\begin{figure}[t]
	\center
	\vspace{-1em}
	\includegraphics[width=\linewidth]{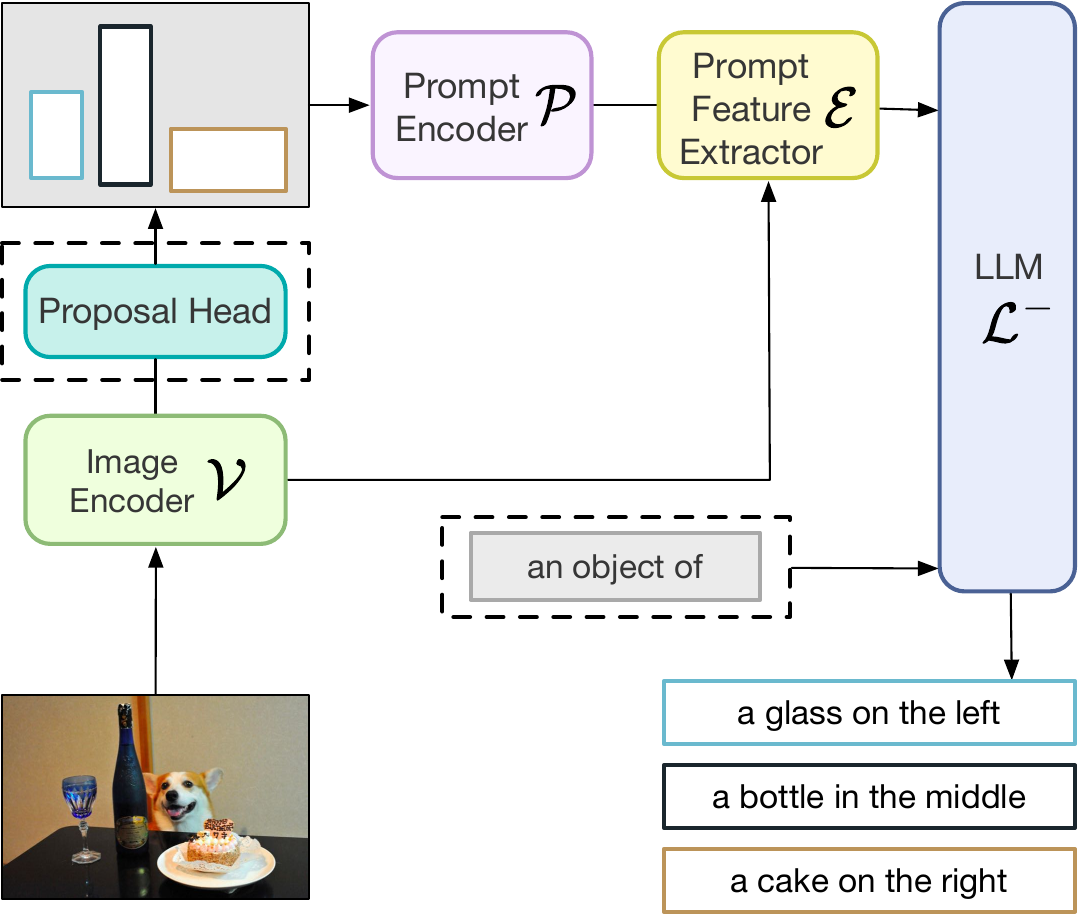}
	\vspace{-1em}
	\caption{
		\label{fig:densecap}
		\textbf{Our model for location-conditioned captioning and dense object captioning.}
  For location-conditioned captioning, the input bounding boxes are provided. For dense object captioning, we first apply a proposal head on the image feature to generate the bounding boxes. We input bounding boxes and image features into the prompt encoder and prompt feature extractor to extract the location-specific feature for each bounding box.  The language model auto-regressively predicts the caption of each object.
	}
	\vspace{-1em}
	\lblfig{fig:dense_cap}
\end{figure}
\noindent\textbf{Location-conditioned captioning} takes an image $\mathbf{I}$ and a bounding box $b$ as a location prompt, and produces a caption sentence $\mathbf{s}^b$ that corresponds to the indicated object from the box query.
Our model can be directly applied to this task, using our prompt encoder and auto-regressive language model, while ignoring the per-word location output.
\begin{equation}
\mathbf{s}^b_i = \mathcal{L}(\mathcal{E}(\mathcal{V}(\mathbf{I}), \mathcal{P}(b)), \mathbf{s}^b_{1:i-1})
\lbleq{referring}
\end{equation}

\begin{figure*}[!t]
	\center
	\vspace{-1em}
	\includegraphics[width=\linewidth]{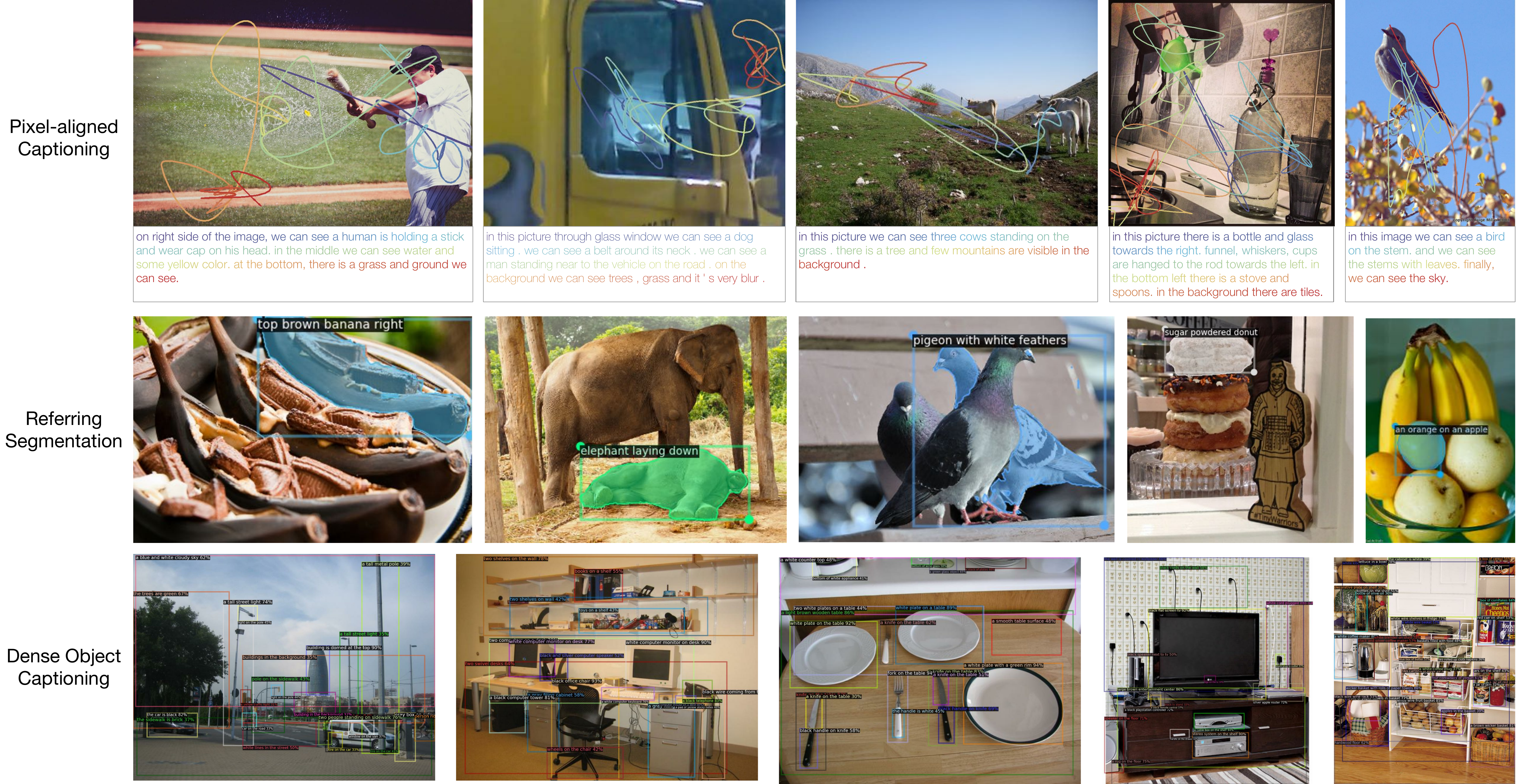}
	\vspace{-1em}
	\caption{
		\label{fig:viz}
		Qualitative results on pixel-aligned captioning (row 1), referring segmentation (row 2), and dense object captioning (row 3). 
  The generated trace semantically corresponds to the caption, represented by color gradient \protect\includegraphics[width=8mm]{figures/gradient.pdf}. 
  In referring segmentation, our model correctly understands the descriptive referring expressions, \eg ``sugar powdered'', ``with white feathers''. 
  For dense object captioning, our model could generate the region-level caption that captures the spatial relationship, \eg ``shelves on the wall''. 
  Zoom in for the best view.
        }
	\vspace{-1em}
\end{figure*}

\noindent\textbf{Dense object captioning} aims to first detect all objects in a given image, and then caption them.
Our framework does not by default detect objects.
To get bounding box candidates, we add a proposal head after our image encoder.
We then feed the resulting bounding boxes separately to the location prompt, to perform location-conditioned captioning for each.
Specifically, we use a Simple Feature Pyramid~\cite{li2022exploring} to upsample the visual features $\mathbf{f}$ to a pyramid of features, and use a CenterNet~\cite{zhou2019objects} head for detection.
We fine-tune the model with the detection losses and the caption losses together end-to-end.

\section{Experiments}

We first introduce our implementation details. 
Then we compare our results against the state of the art on various downstream vision tasks. 
Lastly, we analyze the effectiveness of our proposed components.

\subsection{Implementation details}
\noindent\textbf{Architecture.}
Our visual encoder $\mathcal{V}$ consists of two parallel backbones: ViT-H initialized with SAM~\cite{kirillov2023segment} and ViT-L initialized with EVA02~\cite{fang2023eva}.
The reason for using two backbones is twofold.
First, using the SAM backbone and keeping it frozen enables us to inherit the strong segmentation ability from SAM,
and provide strong localization features for our model.
Second, while keeping the SAM backbone frozen, we use another tunable backbone to learn semantic features.
We concatenate the two features on channel dimension when feeding to the prompt feature extractor.

For the language model $\mathcal{L}$, we use the instruction fine-tuned T5-XL~\cite{2020t5, chung2022scaling} following BLIP2~\cite{li2023blip}. 
We apply low-rank adaption~(LoRA)~\cite{hu2021lora} of rank 32 on query and value projection layers of self-attention blocks to adapt the T5-XL to the vision tasks.
The other parameters of T5-XL are frozen during training.
The prompt feature extractor is a 2-layer transformer with 32 learnable tokens. 
Following PrefixLM paradigm~\cite{2020t5}, we concatenate the prompt feature extractor outputs with the text embedding of the prefix text, and feed it to the language model $\mathcal{L}$.
To generate the segmentation mask, we directly employ the mask decoder of SAM~\cite{kirillov2023segment} on the prompt embedding of the predicted bounding box.
We also include a learnable embedding to enable finetuning the mask on downstream datasets.

\noindent\textbf{Training details.}
We first pretrain our model on WebLI dataset\cite{chen2022pali} with the image captioning objective only, to initialize the prompt feature extractor weights following BLIP2~\cite{li2023blip}. 
During this pretraining, we follow BLIP2(T5-XL)~\cite{li2023blip} to randomly split the image caption into the prefix and the suffix
We concatenate the prefix text with visual features and input them into the encoder of T5-XL.
The suffix text is used as the generation target for the T5-XL decoder. 
We run the pretraining for 10 epochs, using a standard~\cite{li2023blip} captioning input size $224\!\times\!224$.

\begin{table*}[h]
    \tablestyle{10pt}{1.1}
    \begin{tabular}{l|ccc|ccc|cc}
                                        & \multicolumn{3}{c|}{RefCOCO} & \multicolumn{3}{c|}{RefCOCO+} & \multicolumn{2}{c}{RefCOCOg} \\
    Models                             & val     & testA   & testB   & val     & testA    & testB   & val           & test         \\
    \shline
                                        \multicolumn{9}{c}{bounding box P@0.5}                                                    \\
    \hline
TransVG\cite{deng2021transvg}      & 81.0          & 82.7          & 78.4          & 64.8          & 70.7          & 56.9          & 68.7          & 67.7          \\
UNITER\cite{chen2020uniter}        & 81.4          & 87.0          & 74.2          & 75.9          & 81.5          & 66.7          & 74.0          & 68.7          \\
OFA\cite{wang2022ofa}              & 80.0          & 83.7          & 76.4          & 68.3          & 76.0          & 61.8          & 67.6          & 67.6          \\
UniTAB\cite{yang2022unitab}        & 86.3          & 88.8          & 80.6          & 78.7          & 83.2          & 69.5          & 80.0          & 80.0          \\
MDETR\cite{kamath2021mdetr}        & 86.8          & 89.6          & 81.4          & 79.5          & 84.1          & 70.6          & 81.6          & 80.9          \\
VisionLLM\cite{wang2023visionllm}  & 86.7          &               & -             & -             & -             & -             & -             & -             \\
Shikra-7B\cite{chen2023shikra}        & 87.0          & 90.6          & 80.2          & 81.6          & \textbf{87.4} & 71.1          & 82.3          & 82.2          \\
Ferret-7B\cite{you2023ferret}         & 87.5          & 91.4          & 82.5          & 80.8          & \textbf{87.4} & 73.1          & 83.9          & 84.8          \\
\ourmethod{}~(Ours)                 & \textbf{89.8} & \textbf{92.2} & \textbf{86.4} & \textbf{83.2} & 87.0          & \textbf{78.9} & \textbf{84.6} & \textbf{86.0} \\
    \shline
                                        \multicolumn{9}{c}{segmentation mask cIoU}                                                \\
    \hline
ReLA\cite{liu2023gres}             & 73.8          & 76.5          & 70.2          & 66.0          & 71.0          & 57.7          & 65.0          & 66.0          \\
X-Decoder\cite{zou2023generalized} & -             & -             & -             & -             & -             & -             & 64.6          & -             \\
SEEM\cite{zou2023segment}          & -             & -             & -             & -             & -             & -             & 65.7          & -             \\
LISA\cite{lai2023lisa}             & 74.9          & \textbf{79.1} & 72.3          & 65.1          & 70.8          & 58.1          & 67.9          & 70.6          \\
PixelLLM (Ours)                     & \textbf{76.9} & 78.5          & \textbf{74.4} & \textbf{69.2} & \textbf{72.1} & \textbf{64.5} & \textbf{70.7} & \textbf{72.4}
    \end{tabular}
    \vspace{-.5em}
    \caption{
        \label{tab:sota_referring}
        \textbf{State-of-the-art comparison on referring localization and segmentation on RefCOCO.}
        We report the official evaluation metrics: precision at IoU threshold $0.5$ for referring (box) localization, and mask cIoU for referring segmentation.
        Numbers of other methods are taken from the original publications.
        Our model achieves state-of-the-art on all settings for both box and mask outputs.
    }
    \vspace{-1.5em}
\end{table*}

We then train on the Localized Narrative~\cite{pont2020connecting} dataset with the joint captioning and localization objectives in \refeq{loss}.
The default caption in LN contains multiple sentences covering multiple objects in the image. %
We consider each sentence (split by period) as one training target, rather than concatenating all sentences together.
We train for 5 epochs using an input size of $384\!\times\!384$.
The parameters of the prompt encoder, SAM, mask decoder, and T5-XL are frozen during all training stages.
In all settings, we set the localization weight $\lambda$ to 0.1.
We use a label-smooth factor of $0.1$ for our captioning loss.
We provide further details on datasets in the supplement.

\subsection{Joint Captioning and trace generation}

We first evaluate our model trained on the Localized Narrative~\cite{pont2020connecting} dataset, on its validation split. 
We first show qualitative results on joint captioning and trace generation on the top row of \reffig{viz}.
We can see that the generated trace semantically corresponds well to the generated caption.
We next quantitatively evaluate our model under the controlled trace generation setting on the COCO subset of Localized Narrative and compare to MITR~\cite{meng2021connecting},
using their Local Bipartite Matching score (LBM) as the evaluation metric (lower better).
We refer to the MITR paper for details about the metric.
\ourmethod{} achieves $0.153$ LBM (k=0), outperforms $0.163$ of MITR~\cite{meng2021connecting}.
This confirms our model's ability of trace generation task compared to existing works.
We also highlight that our trace outputs are produced from language model features, rather than visual features as in MITR~\cite{meng2021connecting}.

\subsection{SOTA Comparison on Downstream Tasks}

We next fine-tune our model pretrained on Localized Narratives~\cite{pont2020connecting} above on downstream tasks and compare with the state-of-the-art models.

\subsubsection{Referring Localization and Segmentation}
As introduced in \refsec{tasks}, referring localization aims to produce a bounding box or a segmentation mask given a query text.
Following common practice in RefCOCO datasets~\cite{kamath2021mdetr},
we use the combination of RefCOCOs~\cite{yu2016modeling, lin2014microsoft, mao2016generation} and GoldG~\cite{kamath2021mdetr} as the training set for referring expression localization and segmentation.
Note each bounding box (segmentation mask) is annotated with several referring expressions,
and we train on all of them using a regression loss on our bounding box prediction (\refeq{location-prediction}).

We report the official evaluation metrics, bounding box P@0.5 (precision at 0.5 IoU threshold), and segmentation cIoU (cumulative IoU).
\reftbl{sota_referring} shows the results on validation and test split of RefCOCO\cite{lin2014microsoft}, RefCOCO+\cite{yu2016modeling} and RefCOCOg\cite{mao2016generation}.
Our \ourmethod{} outperforms previous and concurrent works \cite{yang2022unitab,kamath2021mdetr,zou2023segment,zou2023generalized, wang2023visionllm,chen2023shikra,you2023ferret,lai2023lisa} on almost all splits of three datasets: 2.3 P@0.5 and 1.4 cIoU on RefCOCO val, 1.7 P@0.5 and 1.5 cIoU on RefCOCO+ val, versus the second best.
We compare to 7B models of~\cite{chen2023shikra, you2023ferret}, which are in comparable size with ours (4B).
While this is a system-level comparison, as models use different backbones and pretraining,
it justifies our model's capability generalizes to this tasks, as our model is not designed specifically for referring. 

\begin{table}[t]
    \tablestyle{10pt}{1.1}
    \begin{tabular}{l|c}
    Model&mAP\\ 
    \shline
    FCLN~\cite{johnson2016densecap}&5.39\\
    JIVC~\cite{yang2017dense}&9.31\\
    ImgG~\cite{li2019learning}&9.25\\
    COCD~\cite{li2019learning}&9.36\\
    COCG~\cite{li2019learning}&9.82\\
    CAG-Net~\cite{yin2019context}&10.51\\
    TDC+ROCSU~\cite{shao2022region}&11.49\\
    GRiT~\cite{wu2022grit}&15.48\\
    \hline
    PixelLLM (Ours)& {\bf 17.02} \\
    \end{tabular}
    \vspace{-.5em}
    \caption{\textbf{State-of-the-art comparison on dense object captioning on Visual Genome}~\cite{krishna2017visual}. We report the official evaluation metric mAP.
    Our model performs the best among all methods.
    \label{tab:sota_densecap}}
    \vspace{-1.5em}
\end{table}
\vspace{-.5em}
\subsubsection{Dense Object Captioning}
\lblsec{densecap}
Dense object captioning aims to detect and caption objects in an image.
As introduced in \refsec{tasks}, our model adds a proposal head on top of our visual features.
The proposal head is randomly initialized and finetuned for this data.
We train and evaluate dense object captioning on Visual Genome~\cite{krishna2017visual} dataset.
We report the official evaluation metric, mAP in \reftbl{sota_densecap}.
In this setting, we use the same proposal network~\cite{zhou2019objects} and language model~\cite{wang2022git} with GRiT~\cite{wu2022grit}.
A key difference between our architecture and GRiT~\cite{wu2022grit} is the region feature extractor, where GRiT uses ROIAlign~\cite{he2017mask}, to explicitly extract image features within the bounding boxes.
\ourmethod{} employs a prompt encoder and prompt feature extractor which don't require the image feature sampling and interpolation like ROIAlign.
Our \ourmethod{} outperforms GRiT by 1.5 mAP,
this again shows the generalization ability of our architecture and pertaining as well as the effectiveness of our prompt feature extractor.

\begin{table}[t]
    \tablestyle{5pt}{1.1}
    \begin{tabular}{l|cc|cc}
                                    & \multicolumn{2}{c|}{RefCOCOg} & \multicolumn{2}{c}{Visual Genome} \\
    Model                           & METEOR        & CIDEr        & METEOR           & CIDEr          \\
                                    \shline
    GRiT~\cite{wu2022grit}          & 13.8          & 75.8         & 17.1             & 142.0          \\
    Kosmos-2~\cite{peng2023kosmos}  & 14.1          & 62.3         & -                & -              \\
    GPT4RoI~\cite{zhang2023gpt4roi} & -             & -            & 17.4             & 145.2          \\
    \ourmethod{}                    & {\bf 14.3}          & {\bf 82.3}         & {\bf 19.9}             & {\bf 148.9}         
    \end{tabular}
    \vspace{-.5em}
    \caption{\textbf{State-of-the-art comparison on Location-conditioned Captioning on RefCOCOg~\cite{mao2016generation} and Visual Genome~\cite{krishna2017visual}.}
    We report averaged per-object METEOR and CIDEr on each dataset.
    Our model outperforms both our GRiT~\cite{wu2022grit} baseline and recent state-of-the-art models.
    \label{tab:sota_loc_cond}}
    \vspace{-.5em}
\end{table}

\subsubsection{Location Conditioned Captioning}
Finally, we evaluate location-conditioned captioning on RefCOCOg\cite{mao2016generation} and Visual Genome\cite{krishna2017visual}.
The input is a ground truth bounding box, and the output is the caption corresponding to the indicated region.
The evaluation metrics are per-box METEOR and CIDEr, similar to standard image captioning~\cite{chen2015microsoft}.
We reuse our dense object captioning model in~\refsec{densecap} and drop the proposal head.

We compare to Kosmos-2\cite{peng2023kosmos} and GPT4RoI~\cite{zhang2023gpt4roi}, who reported results on this task.
We also report results of GRiT~\cite{wu2022grit} evaluated by ourselves.
Kosmos-2\cite{peng2023kosmos} encodes the bounding box coordinates in the text prefix and generates the caption conditioned on it. 
while our \ourmethod{} explicitly encodes the bounding box coordinates using our prompt encoder.
\reftbl{sota_loc_cond} shows the results.
Our \ourmethod{} outperforms previous works, where the improvements can come from our location formulation as well as the localized narrative pretraining.

\subsection{Ablation study}

\begin{table}[t]
\tablestyle{3pt}{1.1}
    \begin{tabular}{@{}l|cc@{}}
Localization Method                                 & Pretrain data  & RefCOCO P@0.5 \\
    \shline
Raw string (\cite{you2023ferret, chen2023shikra})   & GoldG             & 84.4    \\
Extra token (\cite{yang2022unitab, peng2023kosmos}) & GoldG             & 86.1    \\
Regress. on token features                              & GoldG             & 87.6    \\
Regress. on token features (ours)                              & LN+GoldG          & 89.8   
\end{tabular}
\vspace{-.5em}
\caption{\textbf{Ablation on Localization Supervision}
We compare to alternative localization methods including using raw string~\cite{you2023ferret, chen2023shikra} and using extra tokens~\cite{yang2022unitab, peng2023kosmos}.
Both alternatives could only use the sparsely labeled data GoldG~\cite{kamath2021mdetr}, and not the densely labeled data Localized Narratives (LN)~\cite{pont2020connecting}.
Our regression formulation outperforms the alternatives, and can benefit more by using the dense supervision~\cite{pont2020connecting}.
\label{tab:ablation_localization}
}
\vspace{-5mm}
\lbltbl{ablation_localization}
\end{table}
\noindent\textbf{Importance of per-token localization.}
\label{ablation:per_token_localization}
Next, we ablate the importance of our per-token regression formulation for localization in LLMs.
We compare two existing alternatives:
(1) encode the bounding box coordinates as raw strings, and directly use the original LLM output head~\cite{you2023ferret,chen2023shikra}.
(2) discretize the bounding box coordinates into bins following Pix2Seq~\cite{chen2021pix2seq}, and encode the bin number as extra tokens in LLM's vocabulary~\cite{yang2022unitab,peng2023kosmos}.
We note both these two ideas are used to sparsely encode locations, i.e., encode locations for nouns.
Otherwise, if we apply these ideas densely for all words, the output sequence length would be $5\!\times$ longer,
which is not easily feasible to train well.
Our regression formulation allows us to train both sparse localization data~\cite{kamath2021mdetr} (supervise selected words) or dense data~\cite{pont2020connecting} (supervise location outputs for all words).

\reftbl{ablation_localization} shows the results.
First, with all sparse supervision, our formulation (3rd row) slightly improves the two alternatives ($+\!1.5$ points).
The improvement could be from a better localization decoding process:
the alternatives mix decoding regular words and localizations, which the two tasks can interfere with each other.
In our model localization is decoded in parallel.
Second, our formulation enables us to utilize the densely annotated Localized Narratives~\cite{pont2020connecting} as additional training data (4th row).
This further improves our performance by $2.3$ points.
This convincingly shows that our dense word-pixel alignment training objective is beneficial for referring localization tasks.

\begin{table}[!t]
\tablestyle{5pt}{1.1}
    \begin{tabular}{c|c|c|c}
    Language Model & Params & w/ LoRA & RefCOCO P@0.5 \\
\shline
    T5 Small       & 80M    &   \xmark  & 67.0            \\
    T5 Base        & 250M   &   \xmark  & 70.3          \\
    T5 Large       & 780M   &   \xmark  & 73.6          \\
    T5 XL          & 3B     &   \xmark  & 81.9          \\
    \hline
    \hline
    T5 Small       & 80M    & \cmark    & 75.6          \\
    T5 Base        & 250M   & \cmark    & 80.8          \\
    T5 Large       & 780M   & \cmark    & 84.8          \\
    T5 XL          & 3B     & \cmark    & 89.8         
    \end{tabular}
    \vspace{-.5em}
\caption{\textbf{Ablation on language model size and LoRA.}
We report RefCOCO official metrics under different language model sizes.
We observe the localization performance increases consistently with language model size, and using LoRA finetuning is better than freezing the language models.
\label{tab:ablation_lm_size}
}
\vspace{-5mm}
\end{table}
\noindent\textbf{Model size and LoRA fine-tuning.}
We ablate the language model size and LoRA fine-tuning. 
We report the P@0.5 on RefCOCO validation set in Tab.~\ref{tab:ablation_lm_size}.
As we increase the size of the language model, the accuracy on RefCOCO improves consistently both with and without LoRA fine-tuning.
This is expected, as larger models encapsulate more knowledge, which yields stronger understanding and adapting ability.

We note that the T5 models with LoRA~\cite{hu2021lora} outperform the without LoRA counterparts by 8-10 points.
This indicates that LoRA could help the frozen language model adapt to localization tasks better.
It is also worth noting that even without LoRA, the frozen T5-XL performs on par with models that fine-tune the text encoder jointly~\cite{wang2022ofa,chen2020uniter}. 
It is evidence that the frozen large language model like T5 encompasses strong localization ability, which could be revealed by our \ourmethod{}.

\lblsec{results}
\vspace{-2mm}
\section{Conclusion}
\lblsec{conclusion}
We proposed a vision-language model, \ourmethod{}, which can take an image and any combination of location or text as input or output.
\ourmethod{} generates captions, and aligns each output word to a pixel location.
Our model achieved state-of-the-art performance on referring localization, dense captioning, and conditioned captioning.
We hope our work can inspire follow-up works on connecting vision and language, and building general intelligent vision systems.

{\small
\noindent \textbf{Acknowledgements.}
We thank David Ross and Bo Hu for their valuable feedback, and Xinyu Zhang for the help on figures.
The project was supported, in part, by NSF CAREER Award IIS-2240014, Amazon Research Award, Intel Rising Star Faculty Award, and Qualcomm Innovation Fellowship.
}
\begin{spacing}{0.98}
{
    \small
    \bibliographystyle{ieeenat_fullname}
    \bibliography{main}
}
\end{spacing}
\appendix
\section{Dataset details}

\noindent\textbf{Localized Narratives~\cite{pont2020connecting}.}
We use the COCO subset of the Localized Narratives~\cite{pont2020connecting} for both training and evaluation.
It consists of $134,272$ training images and $8,573$ validation images. 
$5,000$ images are annotated with $5$ captions and $5$ traces per image, and the rest are annotated with $1$ caption and $1$ trace per image. We use all the traces for evaluation of controlled trace generation when comparing with MITR~\cite{meng2021connecting}

\vspace{1mm}
\noindent\textbf{GoldG~\cite{kamath2021mdetr}.}
We use the GoldG dataset prepared in MDETR~\cite{kamath2021mdetr} for referring localization. It which consists of images from COCO~\cite{lin2014microsoft}, Visual Genome~\cite{krishna2017visual}, and Flickr30k~\cite{plummer2015flickr30k}. We filtered out all the validation and testing images of RefCOCO, RefCOCO+, and RefCOCOg from our combined training set, yielding $160,280$ training images in total.

\vspace{1mm}
\noindent\textbf{Visual Genome~\cite{krishna2017visual}.}
We use the Visual Genome split prepared in GRiT~\cite{wu2022grit} for dense object captioning, with $77,396$ training images and $5,000$ test images.

\begin{figure}[!t]
	\center
	\vspace{-1em}
	\includegraphics[width=\linewidth]{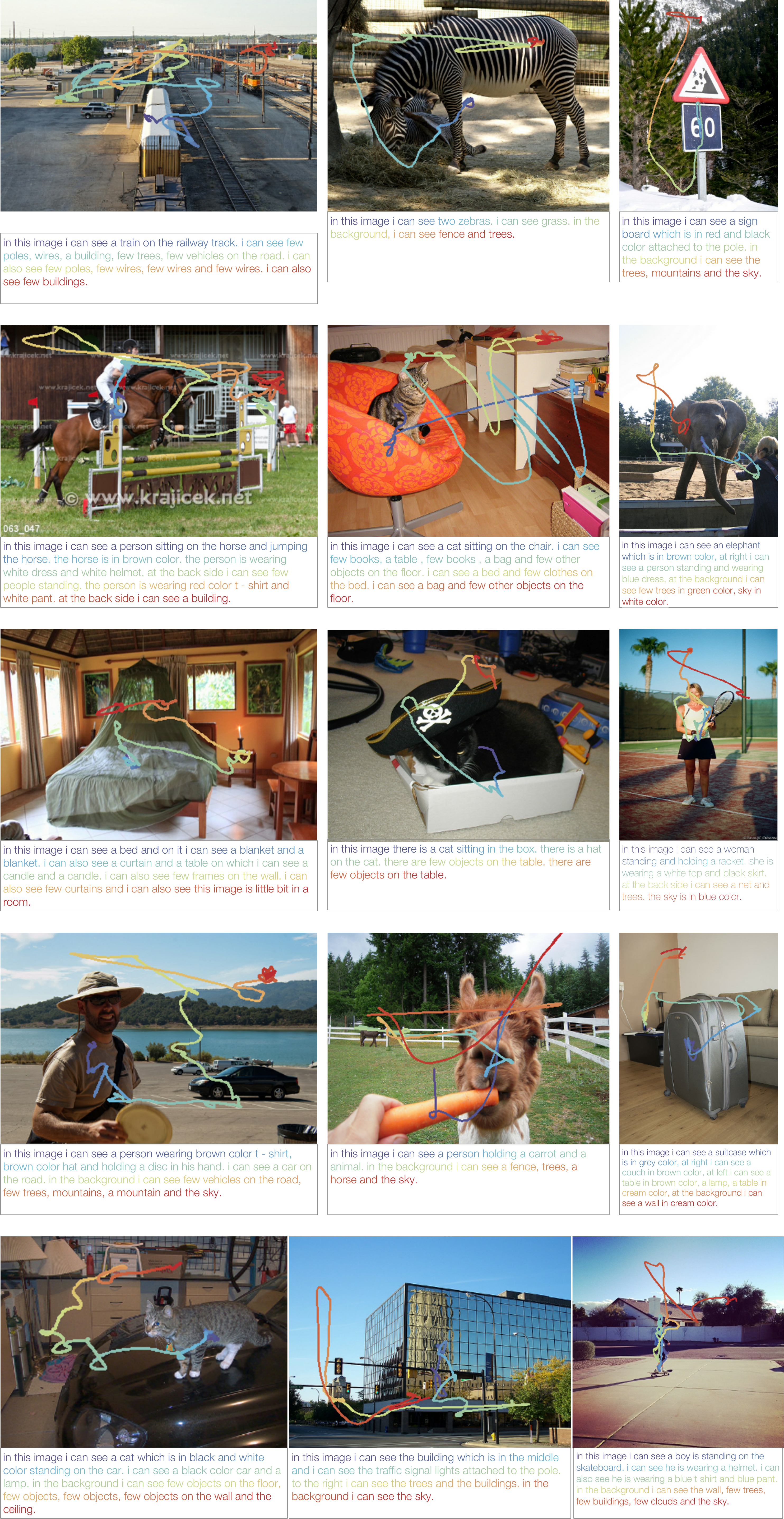}
	\vspace{-1em}
	\caption{
		\label{fig:supp_trc}
		Qualitative results on pixel-aligned captioning.  Zoom in for the best view.
        }
 \vspace{50mm}
\end{figure}

\vspace{-2mm}
\section{Qualitative results}
\begin{figure*}[t]
	\center
	\vspace{-1em}
	\includegraphics[width=\linewidth]{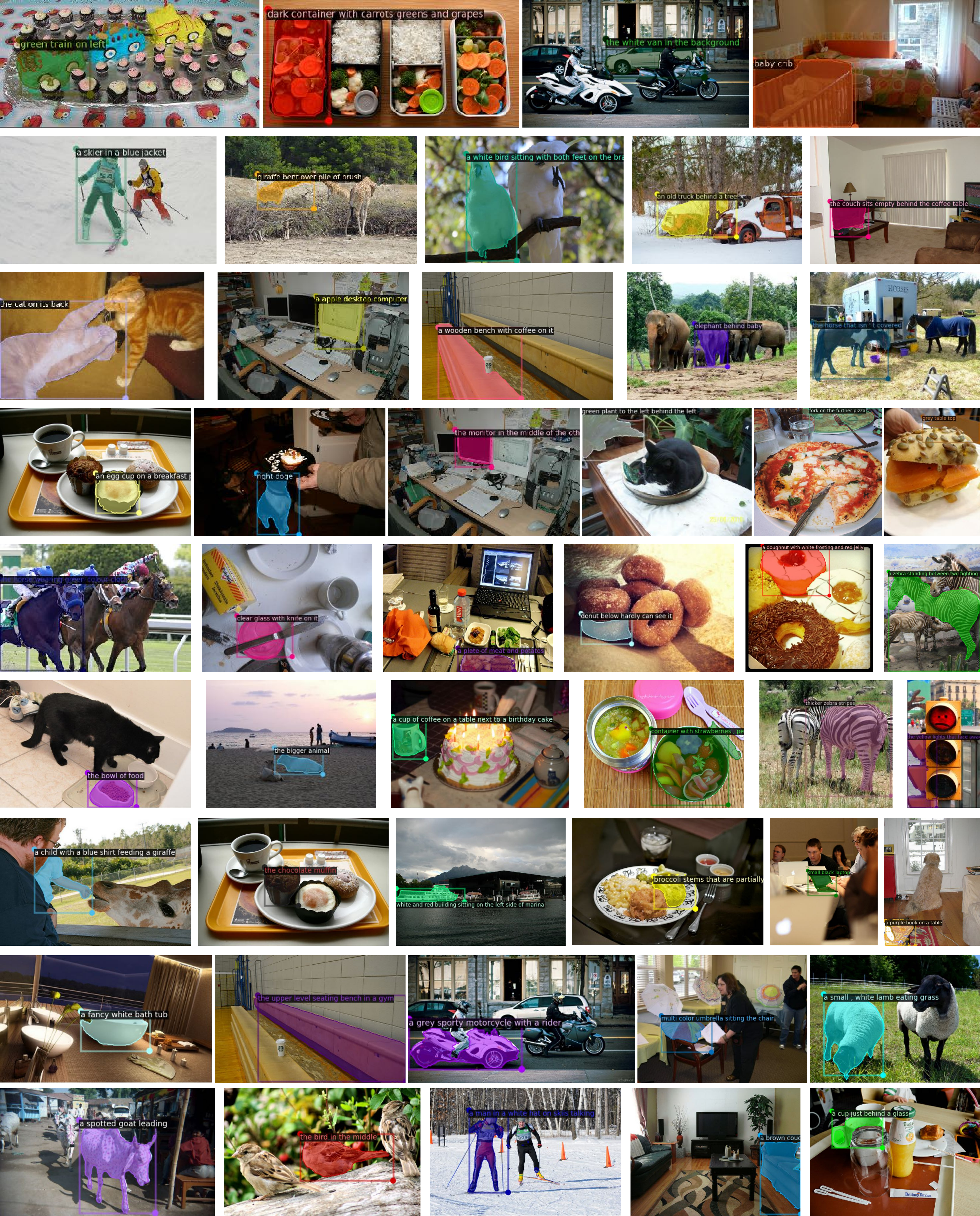}
	\vspace{-1em}
	\caption{
		\label{fig:supp_refm}
		Qualitative results on referring segmentation.  Zoom in for the best view.
        }
	\vspace{-1em}
\end{figure*}

\begin{figure*}[t]
	\center
	\vspace{-1em}
	\includegraphics[width=\linewidth]{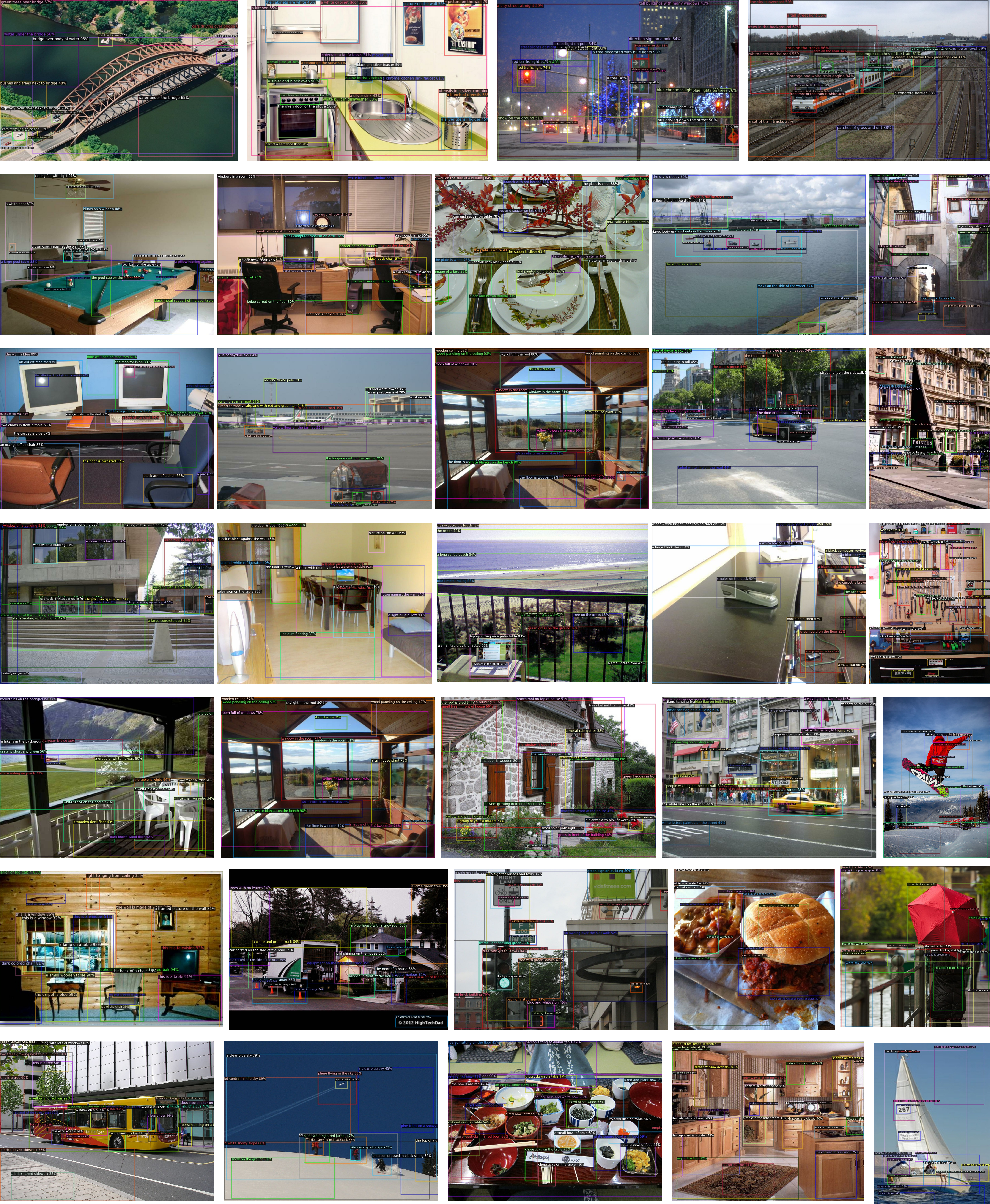}
	\vspace{-1em}
	\caption{
		\label{fig:supp_densecap}
		Qualitative results on dense object captioning.  Zoom in for the best view.
        }
	\vspace{-1em}
\end{figure*}

We provide qualitative more results on pixel-aligned captioning, referring localization and segmentation, and dense object captioning in Figure~\ref{fig:supp_trc},~\ref{fig:supp_refm}, and~\ref{fig:supp_densecap}, respectively.

\end{document}